\documentclass[sigconf]{acmart}

\def\eg{\emph{e.g.}, } 
\def\ie{\emph{i.e.}, }
 
\def\etc{\emph{etc.} }

\definecolor{mygray}{gray}{.9}
\AtBeginDocument{%
  }
\usepackage{multirow}
\usepackage{colortbl}
\usepackage{adjustbox}

\renewcommand\footnotetextcopyrightpermission[1]{}

\makeatletter
\def\authornotetext#1{
 \g@addto@macro\@authornotes{%
 \stepcounter{footnote}\footnotetext{#1}}%
}
\makeatother

\acmConference[MM '24]{Proceedings of the 32nd
ACM International Conference on Multimedia}{October 28-November 1,
2024}{Melbourne, VIC, Australia}

\begin{document}

\title{LDA-AQU: Adaptive Query-guided Upsampling via Local Deformable Attention}

\author[Zewen Du, Zhenjiang Hu, Guiyu Zhao, Ying Jin, and Hongbin Ma]{
	Zewen Du\textsuperscript{\rm 1}, 
	Zhenjiang Hu\textsuperscript{\rm 1}, 
	Guiyu Zhao\textsuperscript{\rm 1}, 
	Ying Jin\textsuperscript{\rm 1},
	Hongbin Ma\textsuperscript{\rm 1,}*
}
\authornotetext{Corresponding author}

\affiliation{
	\textsuperscript{\rm 1} Beijing Institute of Technology, Beijing
	\country{China}
}

\email{{dzw1114, hzj_sixsixsix}@163.com}
\email{{3120220906, jinyinghappy, mathmhb}@bit.edu.cn}

\renewcommand{\shortauthors}{Zewen Du, Zhenjiang Hu, Guiyu Zhao, Ying Jin, $\&$ Hongbin Ma}

\begin{abstract}
  Feature upsampling is an essential operation in constructing deep convolutional neural networks. However, existing upsamplers either lack specific feature guidance or necessitate the utilization of high-resolution feature maps, resulting in a loss of performance and flexibility. In this paper, we find that the local self-attention naturally has the feature guidance capability, and its computational paradigm aligns closely with the essence of feature upsampling (\ie feature reassembly of neighboring points). Therefore, we introduce local self-attention into the upsampling task and demonstrate that the majority of existing upsamplers can be regarded as special cases of upsamplers based on local self-attention. Considering the potential semantic gap between upsampled points and their neighboring points, we further introduce the deformation mechanism into the upsampler based on local self-attention, thereby proposing LDA-AQU. As a novel dynamic kernel-based upsampler, LDA-AQU utilizes the feature of queries to guide the model in adaptively adjusting the position and aggregation weight of neighboring points, thereby meeting the upsampling requirements across various complex scenarios. In addition, LDA-AQU is lightweight and can be easily integrated into various model architectures. We evaluate the effectiveness of LDA-AQU across four dense prediction tasks: object detection, instance segmentation, panoptic segmentation, and semantic segmentation. LDA-AQU consistently outperforms previous state-of-the-art upsamplers, achieving performance enhancements of 1.7 AP, 1.5 AP, 2.0 PQ, and 2.5 mIoU compared to the baseline models in the aforementioned four tasks, respectively. Code is available at \url{https://github.com/duzw9311/LDA-AQU}.
\end{abstract}

%
%



\maketitle

\section{Introduction}

\begin{figure}[!htbp]
  \includegraphics[width=\columnwidth]{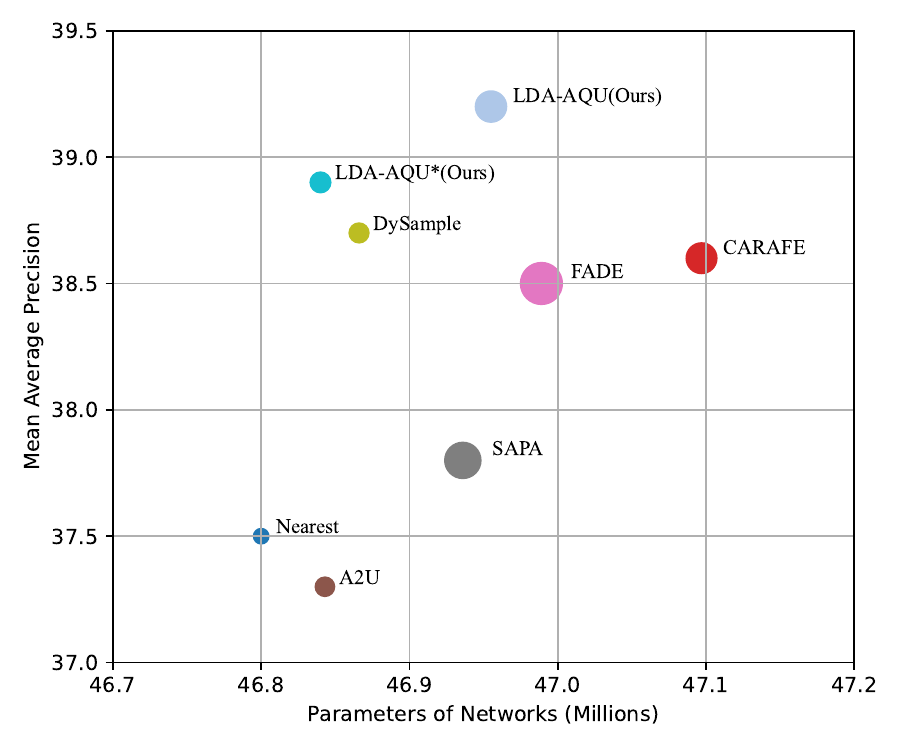}
  \caption{Comparison of various upsamplers in terms of network parameters, Mean Average Precision (mAP) and FLOPs (indicated by area of circles) using Faster R-CNN~\cite{ren2015faster} with ResNet-50~\cite{he2016deep} as the baseline model.}
  \label{fig:comparison}
  \vspace{-0.3cm}
\end{figure}

\begin{figure*}[!t]
  \includegraphics[width=\linewidth]{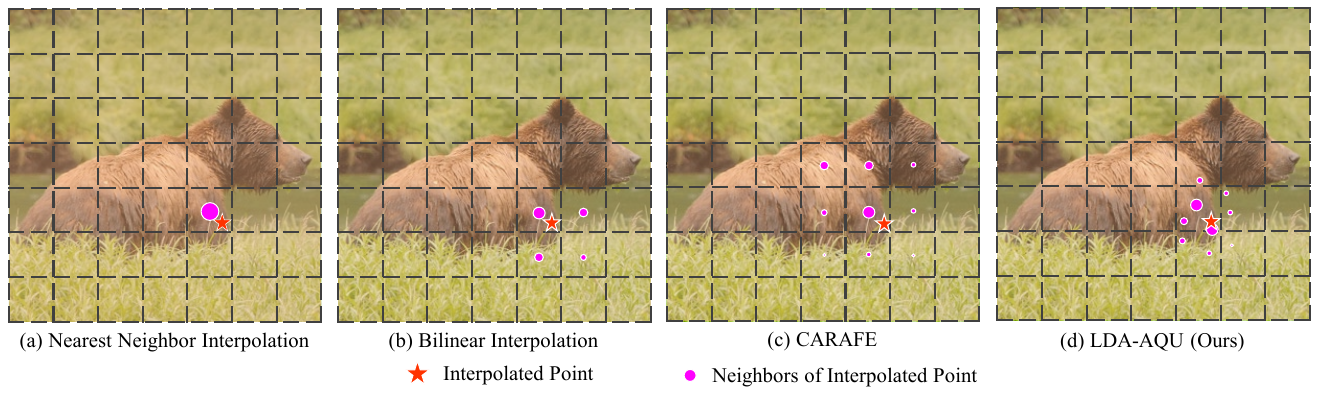}
  \caption{The difference in neighboring point selection schemes between LDA-AQU and other upsamplers including Nearest Neighbor Interpolation, Bilinear Interpolation, and CARAFE~\cite{wang2019carafe}. Given an upsampled point (red star), LDA-AQU employs the query-guided mechanism to predict the deformation offset and aggregation weight of neighboring points, enabling adaptation to upsampling tasks across multiple scales.}
  \label{fig:upsample_dif}
  \vspace{-0.3cm}
\end{figure*}

As a fundamental operator in deep convolutional neural networks, feature upsampling is widely utilized in various dense prediction tasks, including object detection, semantic segmentation, instance segmentation, and image inpainting, \etc 
Given the inherent spatial downsampling characteristics of convolutional and pooling operators, feature upsampling emerges as a crucial inverse operation, indispensable for meeting task-specific requirements. For instance, it facilitates the restoration of spatial resolution in pixel-level dense prediction tasks and enables multi-scale feature fusion within feature pyramid network (FPN)~\cite{lin2017feature}, \etc

Commonly used upsamplers, such as Nearest Neighbor Interpolation and Bilinear Interpolation, aggregate features from neighboring points in a manually designed paradigm, making it difficult to address the requirements of various upsampling tasks simultaneously. Subsequently, several learnable upsampling methods have been proposed, including deconvolution~\cite{long2015fully} and Pixel Shuffle~\cite{shi2016real}, \etc 
However, these methods typically learn a fixed set of parameters for the upsampling kernel, applying the same operation to all spatial positions of the input feature map, leading to suboptimal upsampling results. To enhance the dynamic adaptability of the upsampling operator and enable it to address various upsampling tasks in complex scenes, some upsamplers based on dynamic filters have been proposed and widely applied~\cite{wang2019carafe, zhou2021decoupled, lu2022sapa}. However, these methods either lack specific feature guidance or require the intervention of high-resolution images, limiting their application scenarios and performance.

Most existing upsampling operators can be viewed as a weighted aggregation of features within local neighborhoods surrounding upsampled points (\ie feature reassembly). We observe that this is consistent with the concept of local self-attention, which involves determining an attention weight and then obtaining the necessary contextual information from uniform neighboring points. The difference is that local self-attention naturally incorporates the guidance mechanism (i.e., query guidance), which aligns well with the nature of the upsampling task. This involves adaptively aggregating neighborhood features based on the attributes of the upsampled points, resulting in explicit point affiliations~\cite{lu2022sapa}. These reasons inspire us to explore how to integrate the local self-attention into the upsampling task, aiming to achieve adaptive upsampling with query guidance in a single layer.

In this paper, we introduce a method for incorporating local self-attention into feature upsampling tasks. Additionally, we note that the utilization of fixed and uniform neighboring points may lead to suboptimal upsampling result. As depicted in Figure~\ref{fig:upsample_dif}, uniform neighboring point selection in feature maps with high downsampling strides may result in notable semantic disparities, impeding the generation of high-resolution feature maps. 
Therefore, we further introduce the deformation mechanism to dynamically adjust the positions of neighboring points based on the features of query points (\ie upsampled points) and their contextual information, aiming to further enhance the model's dynamic adaptability. Based on above, we have named our method as LDA-AQU, which offers the following advantages compared to other dynamic upsamplers: 1) operates on a single layer and does not require high-resolution feature maps as input; 2) possesses query-guided capability, enabling the interactive generation of dynamic upsampling kernels using the features of the query points and their neighboring points; 3) exhibits local deformation capability, permitting dynamic adjustment of positions to neighboring points based on the contextual information of query points. These properties enable LDA-AQU to achieve superior performance while remaining lightweight.

Through extensive experiments conducted on four dense predcition tasks including object detection, semantic segmentation, instance segmentation, and panoptic segmentation, we have validated the effectiveness of LDA-AQU. For instance, LDA-AQU can obtain +1.7 AP gains for Faster R-CNN~\cite{ren2015faster}, +1.5 AP gains for Mask R-CNN~\cite{he2017mask}, +2.0 PQ gains for Panoptic FPN~\cite{kirillov2019panopticFPN} on MS COCO dataset~\cite{lin2014microsoft}. On the semantic segmentation task, LDA-AQU also brings +2.5 mIoU gains for UperNet~\cite{xiao2018unified} on ADE20K dataset~\cite{zhou2017scene}. On the aforementioned four dense prediction tasks, our LDA-AQU consistently outperforms the previous state-of-the-art upsampler while maintaining a similar FLOPs and parameters, thereby validating the effectiveness of LDA-AQU.

\section{Related Work}
\subsection{Feature Upsampling}
The most commonly used upsamplers, such as Nearest Neighbor Interpolation and Bilinear Interpolation, are widely adopted in various visual tasks due to their simplicity and efficiency. 
To enhance the adaptability of upsamplers to various tasks, some learnable upsamplers have been proposed. Deconvolution~\cite{long2015fully} employs a reverse convolution operation to achieve the upsampling of feature maps. Pixel Shuffle (PS)~\cite{shi2016real} increases the spatial resolution of the feature map by shuffling features along both spatial and channel directions. With the popularity of dynamic networks~\cite{zhang2020dynet, chen2020dynamic, zhou2021decoupled}, some upsamplers based on dynamic kernels have been introduced. CARAFE~\cite{wang2019carafe} utilizes a content-aware approach to generate dynamic aggregate weights. IndexNet~\cite{lu2019indices} models various upsampling operators as different index functions and proposes several index networks to dynamically generate indexes for guiding feature upsampling. SAPA~\cite{lu2022sapa} introduces the concept of point membership into feature upsampling and guides kernel generation through the similarity between semantic clusters. 
In comparison with them, our LDA-AQU utilizes the features of the query point to generate deformed offsets and aggregate weights of neighboring points in a guided manner, achieving dynamic feature upsampling with feature guidance using only a single low-resolution input.

\subsection{Dense Prediction Tasks}
Dense prediction encompasses pixel-level tasks such as object detection~\cite{girshick2015fast,ren2015faster, redmon2016you, wang2023yolov7, li2020generalized}, instance segmentation~\cite{he2017mask, chen2019hybrid, bolya2019yolact, fang2021instances}, panoptic segmentation~\cite{kirillov2019panoptic, kirillov2019panopticFPN, li2021fully}, and semantic segmentation~\cite{long2015fully, xie2021segformer, cheng2021per}. Models for such tasks typically include a backbone network~\cite{simonyan2014very, he2016deep, howard2017mobilenets}, a feature pyramid network~\cite{lin2017feature, liu2018path, ghiasi2019fpn}, and one or more task heads. The backbone network reduces data dimensions and extracts salient features from input images to decrease computational complexity and capture robust semantic information. The feature pyramid network connects multi-scale features to enhance the model's perception across different scales. The task head links extracted features to the prediction task, serving as the primary distinction among different models.

For instance, Faster R-CNN~\cite{ren2015faster} uses a detection head for object recognition and localization tasks. Mask R-CNN~\cite{he2017mask} achieves both object detection and instance segmentation by adding an instance segmentation head based on Faster R-CNN. Similarly, Panoptic FPN~\cite{kirillov2019panopticFPN} integrates semantic segmentation, instance segmentation, and panoptic segmentation tasks by incorporating an additional semantic segmentation head based on Mask R-CNN. UperNet~\cite{xiao2018unified} uniformly conducts scene perception and parsing by integrating various task heads. Due to the spatial reduction characteristics of convolution and pooling operations, feature upsampling becomes essential for accomplishing above dense prediction tasks. Our LDA-AQU can be easily integrated into these frameworks and consistently brings stable performance improvements.

\subsection{Vision Transformer}
Benefiting from the unique long-range dependency modeling capabilities, Vision Transformer (ViT) and Self-Attention~\cite{dosovitskiy2020image} demonstrate great potential across various visual tasks. As vanilla self-attention capture global contextual information through dense interactions, the computational complexity and memory usage become unbearable for high-resolution inputs. Therefore, in recent years, numerous studies have aimed to optimize the efficiency of ViT. Swin Transformer~\cite{liu2021swin} decreases the number of interactions by confining them to non-overlapping windows, and then captures global dependencies through window sliding. CSwin Transformer~\cite{dong2022cswin} enhances computing efficiency further by employing a cross-shaped interactive window. Additionally, some methods based on local self-attention~\cite{ramachandran2019stand, vaswani2021scaling, zhao2020exploring, pan2023slide} have been proposed to optimize model efficiency and introduce the inductive bias of convolution. We incorporate local self-attention into the upsampling task and introduce a deformation mechanism for neighboring points to enhance the model's dynamic adaptability.

\section{Method}
First, we will provide a brief overview of self-attention and local self-attention. Then, we will elaborate on our approach for extending the local self-attention to address feature upsampling tasks, which we refer to as LA-AQU. Finally, we will introduce the integration of the deformation mechanism into LA-AQU, presenting LDA-AQU as a means to enhance its adaptability in complex scenarios.
\subsection{Preliminary}

\begin{figure}[!t]
  \includegraphics[width=0.7\columnwidth]{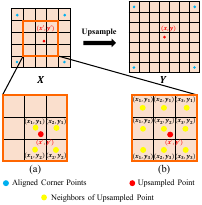}
  \caption{The upsampling kernels of Bilinear Interpolation-Based (a) and Local Self-Attention-Based Upsamplers (b).}
  \label{fig:extern_local}
  \vspace{-0.3cm}
\end{figure}

\subsubsection{Self-Attention}
Given a flattened input feature map $x\in \mathbb{R}^{N \times C}$, where $C$ is the channel size and $N = H \times W$ is the number of tokens along the spatial dimension, the output of the $i$-th token $z_{i}$ after standard self-attention can be expressed as:
\begin{align}
  &(q, k, v) = (xW^{q},xW^{k},xW^{v}) \\ 
  &z_{i} = \sum_{j=1}^{N}\frac{\mbox{exp}(q_{i}k_{j}^\mathrm{T})}{\sum_{m=1}^{N}\mbox{exp}(q_{i}k_{m}^\mathrm{T})}v_{j}
\end{align}
where $W^{q}, W^{k}, W^{v} \in \mathbb{R}^{C \times C}$ are the linear projection matrices. For simplicity, we ignore the output projection matrix $W^{o}$ and the normalization factor $d_{k}$, while also fixing the number of heads to $1$.

The standard self-attention employs dense interaction for each query to gather crucial long-range dependencies. Hence, the computational complexity can be expressed as $\mathcal{O}(2N^{2}C+4NC^{2})$, with $\mathcal{O}(2N^{2}C)$ for dense interaction and $\mathcal{O}(4NC^{2})$ for linear projection.

\begin{figure*}[!bpt]
  \includegraphics[width=\linewidth]{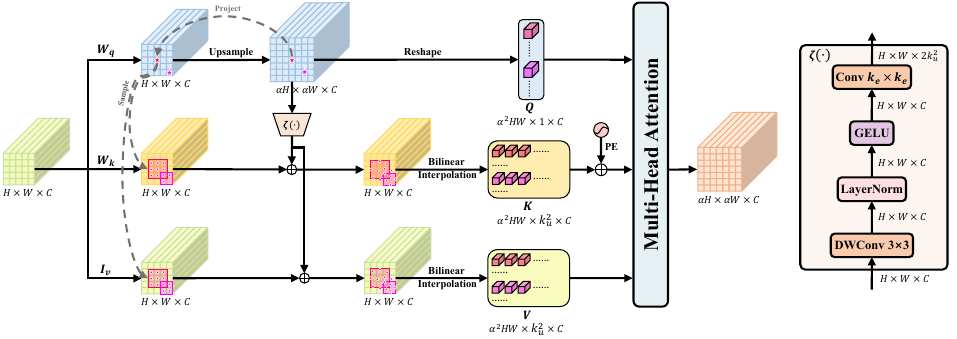}
  \caption{The overall framework of LDA-AQU. Given an input feature map with size $H \times W \times C$ and an upsampling factor $\alpha$, LDA-AQU employs local deformable attention for feature upsampling, resulting in an output feature map with size $\alpha H \times \alpha W \times C$.}
  \label{fig:overal_archi}
  \vspace{-0.3cm}
\end{figure*}
\subsubsection{Local Self-Attention}
Due to the quadratic scaling of computational complexity with the number of tokens, some researchers are exploring the use of local self-attention to replace standard self-attention, aiming to reduce computational complexity and introduce the local induction bias of convolution. Assuming the kernel size of the neighborhood sampling is $n$. For a flattened input feature map $x\in \mathbb{R}^{N \times C}$, the output of the $i$-th token $z_{i}$ after local self-attention can be expressed as:
\begin{align}
  &(q, k, v) = (xW^{q},xW^{k},xW^{v}) \\ 
  &(\tilde{q}, \tilde{k}, \tilde{v}) = (\mbox{Reshape}(q),\phi(k,n),\phi(v,n)) \\
  &z_{i} = \sum_{j=1}^{n^{2}}\frac{\mbox{exp}(\tilde{q}_{i}\tilde{k}_{j}^\mathrm{T})}{\sum_{m=1}^{n^{2}}\mbox{exp}(\tilde{q}_{i}\tilde{k}_{m}^\mathrm{T})}\tilde{v}_{i}
\end{align}
where $\tilde{q} \in \mathbb{R}^{N\times 1 \times C}$ denotes the queries after the $Reshape$ operation. $\phi(\cdot,\cdot)$ is the sampling function of neighborhood points, which can be easily accomplished by the built-in funcion $unfold$ in PyTorch~\cite{paszke2017automatic}. The $\tilde{k}, \tilde{v} \in \mathbb{R}^{N\times n^{2} \times C}$ are the keys and values after neighborhood sampling, respectively.

The local self-attention restricts feature interaction to the local neighborhood of each query, thus the computational complexity can be expressed as $\mathcal{O}(2n^{2}NC+4NC^{2})$, with $\mathcal{O}(2n^{2}NC)$ for feature interaction and $\mathcal{O}(4NC^{2})$ for linear projection.

\subsection{Feature Upsampling via Local Deformable Attention}
\subsubsection{Extending Local Self-Attention for Upsampling}
Given an input feature map $X \in \mathbb{R}^{H \times W \times C}$ and upsampling factor $\alpha \in [1, +\infty]$, the output feature map $Y \in \mathbb{R}^{\alpha H \times \alpha W \times C}$ can be obtained through feature upsampling. 
Initially, the queries, keys, and values of the input feature map can be obtained through the linear mapping.
\begin{equation}
  (Q, K, V) = (XW^{Q}, XW^{K}, XW^{V}) \label{equ:qkv}
\end{equation}

Assuming the kernel size of neighborhood sampling is $k_{u}=3$. As illustrated in Figure~\ref{fig:extern_local}(b), let $p=(x, y)$ denote the coordinate of the point to be interpolated, where $x \in [0, W-1]$ and $y \in [0, H-1]$. Taking the grid arrangement format of aligned corner points as an example, we can obtain the corresponding coordinate $p'=(x', y')$ in the input feature map by
\begin{equation}
  p'= \psi(p) = (x\frac{W}{\alpha W - 1}, y\frac{H}{\alpha H - 1}) \label{equ:proj}
\end{equation}

Let $r=\{(x_{1}, y_{1}), (x_{2}, y_{1}), ..., (x_{3}, y_{3})\}$ denote the absolute coordinates of uniform neighboring points of $p'$. The upsampled result of point $p$ based on the local self-attention can be expressed as:
\begin{equation}
  Y(p) = \sum_{s\in r}\frac{\mbox{exp}(Q(p')K(s)^\mathrm{T})}{\sum_{t\in r}\mbox{exp}(Q(p')K(t)^\mathrm{T})}V(s)
\end{equation}

Let $F(p', s)=\frac{\mbox{exp}(Q(p')K(s)^\mathrm{T})}{\sum_{t\in r}\mbox{exp}(Q(p')K(t)^\mathrm{T})}$. When $W^{V}$ is an identity matrix, the above formula can be further simplified to
\begin{equation}
  Y(p) = \sum_{s\in r}F(p', s)X(s) \label{equ:la_func}
\end{equation}

Considering the properties of $Softmax$, we have $\sum_{s\in r}F(p', s)=1$. 
Therefore, LA-AQU can be described as the adaptive acquisition of aggregation weights guided by the query features and the reassembly of features from neighboring points. Indeed, many existing upsamplers can be formulated by upsampler based on local self-attention. We will delve into this concept in Section~\ref{sec:relating}.

\subsubsection{Introducing Deformation Mechanism}
LA-AQU employs a fixed and uniform neighborhood sampling scheme for feature upsampling. Consequently, it becomes challenging to meet the upsampling needs across various complex scenarios simultaneously. As shown in Figure~\ref{fig:upsample_dif}, solely aggregating the features of uniformly sampled neighboring points will result in the model overly emphasizing less significant background regions while neglecting the features of the object itself. Therefore, we further introduce the deformation mechanism and propose LDA-AQU. 

The overall process of LDA-AQU is illustrated in Figure~\ref{fig:overal_archi}. Similarly, considering the input feature map $X \in \mathbb{R}^{H \times W \times C}$ and the output feature map $Y \in \mathbb{R}^{\alpha H \times \alpha W \times C}$, we initially obtain the $Q$, $K$, $V$ through Equation~\ref{equ:qkv}. To avoid continuous projection and sampling, we employ bilinear interpolation to upsample the matrix $Q\in \mathbb{R}^{H\times W \times C}$, resulting in $Q'\in \mathbb{R}^{\alpha H\times \alpha W \times C}$. Then, utilizing the built-in function $meshgrid$ in PyTorch, we can generate the uniform coordinate matrix $P\in \mathbb{R}^{\alpha H\times \alpha W \times 1\times 2}$ for the upsampled feature map. Through Equation~\ref{equ:proj}, we can derive the reference point coordinate matrix $P'$ by projecting $P$ onto the input feature map.

Assuming the kernel size of neighborhood sampling is $k_{u}$, the initial offset $\varDelta P$ of neighboring points can be obtained as 
\begin{equation}
  \varDelta P=\{(-\lfloor k_{u} / 2 \rfloor, -\lfloor k_{u} / 2\rfloor),..., (\lfloor k_{u} / 2 \rfloor, \lfloor k_{u} / 2 \rfloor)\}
\end{equation}

Subsequently, we can obtain the coordinate matrix of neighboring points $R \in \mathbb{R}^{\alpha H\times \alpha W \times k_{u}^{2}\times 2}$ of the upsampled points through the broadcast mechanism by $R = P' + \varDelta P$.

To enable the neighboring points to dynamically adjust their positions, we introduce a sub-network $\zeta (\cdot )$, utilizing query features to generate the query-guided sampling point offset matrix $\varDelta R \in \mathbb{R}^{\alpha H\times \alpha W \times k_{u}^{2}\times 2}$. 
Based on the uniform neighboring point coordinate matrix $R$ and the predicted offset matrix $\varDelta R$, the final deformed neighboring point coordinate matrix $R'$ can be obtained by
\begin{equation}
  R'=R+\varDelta R =R+\zeta (Q)
\end{equation}

Finally, we utilize the point-wise features of matrix $Q'$ to interact with the corresponding local point features of $K, V$ to complete the upsampling task. The aforementioned process can be expressed as:
\begin{align}
  (\widetilde{Q},\widetilde{K},\widetilde{V}) &= (\mbox{Reshape}(Q'), \Phi (K, R'), \Phi (V, R')) \\
  Y &= \mbox{Softmax}(\frac{\widetilde{Q}\widetilde{K}^\mathrm{T}}{\sqrt{d_{k}}})\widetilde{V}
\end{align}
where $\widetilde{Q} \in \mathbb{R}^{\alpha H \times \alpha W \times 1 \times C}$ denotes the queries after $Reshape$ operation. $\widetilde{K}, \widetilde{V} \in \mathbb{R}^{\alpha H \times \alpha W \times k_{u}^{2} \times C}$ are the keys and values after neighboring sampling. To ensure differentiability, we use bilinear sampling function $\Phi (\cdot, \cdot)$ for sampling with non-integer offsets.

\textbf{Offset Predictor. }
The detailed structure of the offset predictor $\zeta (\cdot)$, utilized to generate the sampling point offset matrix $\varDelta R$, is depicted in Figure~\ref{fig:overal_archi}.
We employ a $3\times 3$ depthwise convolution to extend the perceptual range of the queries and utilize a convolutional layer with the kernel size of $k_{e}$ to predict the offset for $k_{u}^{2}$ points.
The ablation studies of $k_{e}$ and $k_{u}$ can be found in Section~\ref{sec:ablation}.

\textbf{Offset Groups. }
To augment the model's capability in perceiving distinct channel features and to enhance its adaptability across diverse scenarios, we partition the channels of the input feature maps within the offset predictor $\zeta (\cdot)$, employing varied neighboring points for different feature groups.

\textbf{Local Deformation Ranges. }
To expedite convergence speed, we apply the $Tanh$ function to the results of the Offset Predictor, limiting them to the range $[-1, 1]$. Additionally, we employ a factor $\theta $ to regulate the deformation range of the neighboring points. In our experiments, we observe that increasing the value of $\theta $ will bring some performance improvements. 
We believe that employing a larger local deformation range, combined with the grouping operation, will enable the model to capture a wider range of crucial features within the neighborhood for refining the upsampling results. More detailed analysis will be conducted in Section~\ref{sec:ablation}.

\subsection{Relating to Other Upsampling Methods} \label{sec:relating}
In this section, we will explore the relationship between different upsamplers and LDA-AQU. Indeed, LA-AQU can already represent the majority of upsampling schemes.

\textbf{Bilinear Interpolation. }
Similarly, let $X\in \mathbb{R}^{H \times W \times C}$ and $Y\in \mathbb{R}^{\alpha H \times \alpha W \times C}$ denote the input and output feature map, respectively. As shown in Figure~\ref{fig:extern_local}(a), after projection via Equation~\ref{equ:proj}, the feature vector of the upsampled point $p=(x, y)$ can be expressed as 
\begin{equation}
  Y(p)=\sum_{s\in r}F(p', s)X(s) \label{equ:bi_func}
\end{equation}
where $r=\{(x_{1}, y_{1}), (x_{1}, y_{2}), (x_{2}, y_{1}), (x_{2}, y_{2})\}$ is the set of standard neighbor points of point $p'$. 
The bilinear interpolation kernel $F(p', s)$ in here can be represented as
\begin{equation}
  F(p', s)=w(x', s_{x})w(y', s_{y})
\end{equation}
where $w(a, b)=\max(0, 1-|a-b|)$ represents the aggregation weight based on the distance between point pairs.

By comparing Equation~\ref{equ:bi_func} and Equation~\ref{equ:la_func}, we can conclude that upsampling through bilinear interpolation is indeed a special instance of LA-AQU. 
When the aggregate weight of points $\{(x_{1}, y_{1}), (x_{2}, y_{1}), (x_{3}, y_{1}), (x_{1}, y_{2}), (x_{1}, y_{3})\}$ of $F(p', s)$ in Equation~\ref{equ:la_func} equals zero, and the computation results are based on distance, LA-AQU will degrade into upsampling based on bilinear interpolation.

\textbf{Nearest Neighbor Interpolation. }
When only the aggregate weight of the nearest neighboring point of upsampled points is non-zero, LA-AQU will degrade into upsampling based nearest neighbor interpolation. 

\textbf{CARAFE. }
CARAFE and LA-AQU are fundamentally similar in functionality as they both compute dynamic content-aware aggregation weights for neighboring points of upsampled points. The distinction lies in LA-AQU utilizing query-related dynamic features for interactive prediction of kernel weights, whereas CARAFE employs a convolutional alyer for predicting kernel weights. Moreover, the query-guided weight generation in LA-AQU enables the model to prioritize features that closely align with its own content.

\textbf{DySample. }
DySample achieves feature upsampling through sampling. After introducing the deformation mechanism, LDA-AQU can achieve similar effects. However, LDA-AQU employs the query-guided mechanism for neighboring deformation and feature aggregation. Compared with DySample, LDA-AQU can more effectively leverage object-specific information, and the kernel-based upsampling scheme aligns better with human intuition, \ie inferring the features of upsampled points based on their neighbors.

It is noteworthy that LDA-AQU avoids using the PixelShuffle operator, in contrast to CARAFE and DySample, enabling it to achieve any desired multiple of feature upsampling.

\subsection{Complexity Analysis}
Given an input feature map with the shape of $H \times W\times C$, the overall computational complexity of LDA-AQU can be expressed as  $\mathcal{O}(2HWC^{2}+2\alpha ^{2}k_{u}^{2}HWC+2\alpha ^{2}k_{u}^{2}k_{e}^{2}HWC)$, with $\mathcal{O}(2HWC^{2})$ for linear projection, $\mathcal{O}(2\alpha ^{2}k_{u}^{2}HWC)$ for attention interaction, and $\mathcal{O}(2\alpha ^{2}k_{u}^{2}k_{e}^{2}HWC)$ for deformed offsets prediction. Note that we ignore the computational complexity of depthwise convolution and positional encoding since their computational complexity is considerably lower compared to the aforementioned blocks. To summarize, LDA-AQU exhibits linear computational complexity with the number of input tokens (\ie image resolution). 

\begin{table*}[!t]
  \caption{Performance comparison of Object Detection on MS COCO based on Faster R-CNN. * denotes that the channel reduction factor is set to 16 to balance FLOPs and performance. The best is highlighted in bold, and the second best is underlined. }
  \centering
  \begin{tabular}{l|c|ccc|ccc|cc|c}
  \toprule
  Faster R-CNN & Backbone & $AP$ & $AP_{50}$ & $AP_{75}$ & $AP_S$  & $AP_M$  & $AP_{L}$ & Params& FLOPs & Reference\\ 
  \midrule
  Nearest & ResNet-50 & 37.5 & 58.2 & 40.8 & 21.3 & 41.1 & 48.9 & 46.8M & 208.5G & - \\
  Deconv & ResNet-50 & 37.3 & 57.8 & 40.3 & 21.3 & 41.1 & 48.0 & +2.4M & +12.6G & - \\
  PS~\cite{shi2016real} & ResNet-50 & 37.5 & 58.5 & 40.4 & 21.5 & 41.5 & 48.3 & +9.4M & +50.2G & CVPR16 \\
  CARAFE~\cite{wang2019carafe} & ResNet-50 & 38.6 & 59.9 & 42.2 & \textbf{23.3} & 42.2 & 49.7 & +0.3M & +1.6G & ICCV19\\
  IndexNet~\cite{lu2019indices} & ResNet-50 & 37.6 & 58.4 & 40.9 & 21.5 & 41.3 & 49.2 & +8.4M & +46.4G & ICCV19 \\
  A2U~\cite{dai2021learning} & ResNet-50 & 37.3 & 58.7 & 40.0 & 21.7 & 41.1 & 48.5 & +38.9K & +0.3G & CVPR21 \\
  FADE~\cite{lu2022fade} & ResNet-50 & 38.5 & 59.6 & 41.8 & \underline{23.1} & 42.2 & 49.3 & +0.2M & +3.4G & ECCV22 \\
  SAPA-B~\cite{lu2022sapa} & ResNet-50 & 37.8 & 59.2 & 40.6 & 22.4 & 41.4 & 49.1 & +0.1M & +2.4G & NeurIPS22\\
  DySample~\cite{liu2023learning} & ResNet-50 & 38.7 & 60.0 & 42.2 & 22.5 & 42.4 & \textbf{50.2} & +65.5K & +0.3G & ICCV23\\
  \rowcolor{mygray} LDA-AQU* & ResNet-50  & \underline{38.9} & \underline{60.4} & \underline{42.4} & \textbf{23.3} & \underline{42.8} & 49.7 & +41.0K & +0.4G & -\\
  \rowcolor{mygray} LDA-AQU & ResNet-50  & \textbf{39.2} & \textbf{60.7} & \textbf{42.7} & 22.9 & \textbf{43.0} & \underline{50.1} & +0.2M & +1.7G & -\\
  \bottomrule
  \end{tabular}
  \label{tab:faster_rcnn}
\end{table*}
\section{Experiments}

\begin{table}[!htbp] 
  \caption{Performance comparison of Instance Segmentation on MS COCO based on Mask R-CNN.}
  \centering
  \begin{adjustbox}{width=0.45\textwidth}
  \begin{tabular}{l|cc|ccc|ccc}
  \toprule
  Mask R-CNN & Task & Backbone & $AP$ & $AP_{50}$ & $AP_{75}$ & $AP_S$ & $AP_M$ & $AP_{L}$ \\
  \midrule
  Nearest  & Bbox & ResNet-50 & 38.3 & 58.7 & 42.0 & 21.9 & 41.8 & 50.2 \\
  Deconv  &  & ResNet-50 & 37.9 & 58.5 & 41.0 & 22.0 & 41.6 & 49.0 \\
  PS~\cite{shi2016real}  &  & ResNet-50 & 38.5 & 59.4 & 41.9 & 22.0 & 42.3 & 49.8 \\
  CARAFE~\cite{wang2019carafe} & & ResNet-50 & 39.2 & 60.0 & 43.0 & 23.0 & 42.8 & 50.8 \\
  IndexNet~\cite{lu2019indices} & & ResNet-50 & 38.4 & 59.2 & 41.7 & 22.1 & 41.7 & 50.3 \\
  A2U~\cite{dai2021learning} & & ResNet-50 & 38.2 & 59.2 & 41.4 & 22.3 & 41.7 & 49.6 \\
  FADE~\cite{lu2022fade} & & ResNet-50 & 39.1 & 60.3 & 42.4 & \underline{23.6} & 42.3 & 51.0 \\
  SAPA-B~\cite{lu2022sapa} & & ResNet-50 & 38.7 & 59.7 & 42.2 & 23.1 & 41.8 & 49.9\\
  DySample~\cite{liu2023learning} & & ResNet-50 & \underline{39.6} & \underline{60.4} & \textbf{43.5} & 23.4 & \underline{42.9} & \textbf{51.7}\\
  \rowcolor{mygray} LDA-AQU  & & ResNet-50 & \textbf{39.8} & \textbf{60.8} & \textbf{43.5} & \textbf{23.8} & \textbf{43.6} & \underline{51.6}\\
  \midrule
  Nearest & & ResNet-101 & 40.0 & 60.4 & 43.7 & 22.8 & 43.7 & 52.0 \\
  DySample & & ResNet-101 & \underline{41.0} & \underline{61.9} & \underline{44.9} & \underline{24.3} & \underline{45.0} & \underline{53.5}\\
  \rowcolor{mygray} LDA-AQU  & & ResNet-101 & \textbf{41.3} & \textbf{62.3} & \textbf{45.2} & \textbf{24.4} & \textbf{45.5} & \textbf{53.7}\\
  \midrule
  Nearest & Segm & ResNet-50 & 34.7 & 55.8 & 37.2 & 16.1 & 37.3 & 50.8 \\
  Deconv  & & ResNet-50 & 34.5 & 55.5 & 36.8 & 16.4 & 37.0 & 49.5 \\
  PS~\cite{shi2016real} & & ResNet-50 & 34.8 & 56.0 & 37.3 & 16.3 & 37.5 & 50.4 \\
  CARAFE~\cite{wang2019carafe} & & ResNet-50 & 35.4 & 56.7 & 37.6 & \underline{16.9} & 38.1 & 51.3 \\
  IndexNet~\cite{lu2019indices} & & ResNet-50 & 34.7 & 55.9 & 37.1 & 16.0 & 37.0 & 51.1 \\
  A2U~\cite{dai2021learning} & & ResNet-50 & 34.6 & 56.0 & 36.8 & 16.1 & 37.4 & 50.3 \\
  FADE~\cite{lu2022fade} & & ResNet-50 & 35.1 & 56.7 & 37.2 & 16.7 & 37.5 & 51.4 \\
  SAPA-B~\cite{lu2022sapa} & & ResNet-50 & 35.1 & 56.5 & 37.4 & 16.7 & 37.6 & 50.6\\
  DySample~\cite{liu2023learning} & & ResNet-50 & \underline{35.7} & \underline{57.3} & \underline{38.2} & \textbf{17.3} & \underline{38.2} & \underline{51.8}\\
  \rowcolor{mygray} LDA-AQU  & & ResNet-50 & \textbf{36.2} & \textbf{57.9} & \textbf{38.5} & \textbf{17.3} & \textbf{39.1} & \textbf{52.8}\\
  \midrule
  Nearest & & ResNet-101 & 36.0 & 57.6 & 38.5 & 16.5 & 39.3 & 52.2 \\
  DySample & & ResNet-101 & \underline{36.8} & \underline{58.7} & \underline{39.5} & \underline{17.5} & \underline{40.0} & \underline{53.8}\\
  \rowcolor{mygray} LDA-AQU  & & ResNet-101 & \textbf{37.5} & \textbf{59.2} & \textbf{40.2} & \textbf{17.6} & \textbf{41.1} & \textbf{54.3}\\
  \bottomrule
  \end{tabular}
  \end{adjustbox}
  \label{tab:mask_rcnn}
\end{table}
\subsection{Experimental Settings}
We evaluate the effectiveness of proposed LDA-AQU on four challenging tasks, including object detection, instance segmentation, semantic segmentation, and panoptic segmentation. 

\textbf{Datasets and Evaluation Metrics. }
We utilize the challenging MS COCO 2017 dataset~\cite{lin2014microsoft} to evaluate the effectiveness of LDA-AQU across object detection, instance segmentation, and panoptic segmentation tasks, respectively. For object detection and instance segmentaion tasks, we report the standard COCO metrics of Mean Average Precision (mAP). For the panoptic segmentation task, we report the PQ, SQ, and RQ metrics~\cite{kirillov2019panoptic} as in Dysample~\cite{liu2023learning}. For the semantic segmentation task, we conduct performance comparison using the ADE20K dataset~\cite{zhou2017scene} and report the Average Accuracy (aAcc), Mean IoU (mIoU), Mean Accuracy (mAcc) metrics.

\textbf{Implementation Details. }
We evaluate the effectiveness of LDA-AQU using MMDetection~\cite{chen2019mmdetection} and MMSegmentation~\cite{contributors2020mmsegmentation} toolboxes. Specifically, we adopt Faster R-CNN~\cite{ren2015faster}, Mask R-CNN~\cite{he2017mask}, Panoptic FPN~\cite{kirillov2019panopticFPN} and UperNet~\cite{xiao2018unified} as the baseline models. If unspecified, the offset groups, local deformation ranges, and channel size reduction factors are set to 2, 11, and 4, respectively. We use 1$\times$ training schedule for object detection, instance segmentation and panoptic segmentation tasks. For semantic segmentation, the model is trained for 160K iterations. All other training strategies and hyperparameters remain the same as in \cite{liu2023learning} for fair comparison.

\begin{figure*}[!t]
  \includegraphics[width=\linewidth]{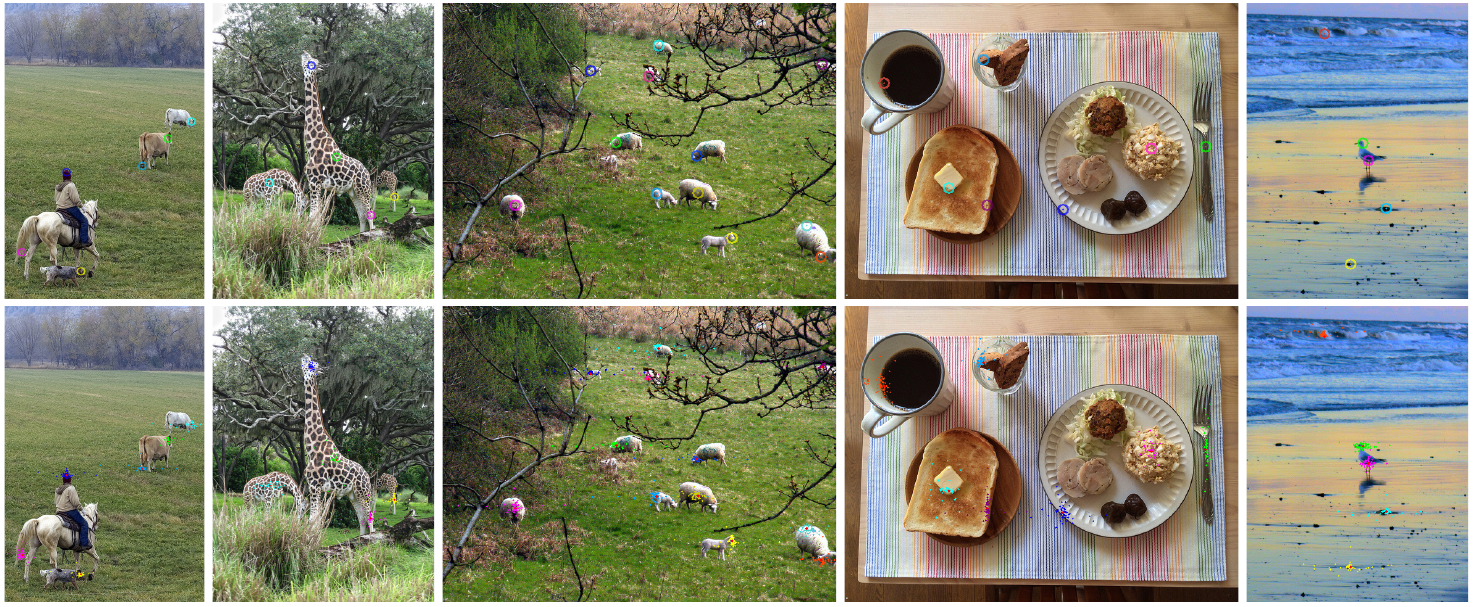}
  \caption{Visualizations of some upsampled points (first row) and their deformed neighboring points (second row). Colored rings depict upsampled points, while scatter points of the same color are corresponding deformed neighboring points.}
  \label{fig:visual_pts}
  \vspace{-0.3cm}
\end{figure*}

\subsection{Object Detection}
As shown in Table~\ref{tab:faster_rcnn}, LDA-AQU outperforms the best model, DySample, by a large margin of 0.5 AP (39.2 AP v.s 38.7 AP). Moreover, LDA-AQU incurs only a minor increase in FLOPs and parameters, nearly matching CARAFE but surpassing it by 0.6 AP. For fair comparison, we adjust the channel size reduction factor to $16$ to ensure a comparable FLOPs and parameters to DySample. Results in Table~\ref{tab:faster_rcnn} indicate that the LDA-AQU maintains superior performance even with similar FLOPs and parameters (38.9 AP v.s 38.7 AP).

\subsection{Instance Segmentation}
As illustrated in Table~\ref{tab:mask_rcnn}, LDA-AQU improves the performance of the Mask RCNN by 1.5 bbox AP (39.8 AP v.s 38.3 AP) and 1.5 mask AP (36.2 AP v.s 34.7 AP), surpassing DySample by 0.2 bbox AP (39.8 AP v.s 39.6 AP) and 0.5 mask AP (36.2 AP v.s 35.7 AP). Considering the size of the input feature map in mask head is $14 \times 14$, we reduce the local deformation range of LDA-AQU embedded into the mask head to $5$ to prevent instability in the training process caused by excessive deformation range. The detailed ablation study of the deformation ranges on the mask head can be found in Section~\ref{sec:ablation}.

\subsection{Panoptic Segmentation}
As shown in Table~\ref{tab:panoptic_fpn}, LDA-AQU surpasses all previous methods by a significant margin and maintains a similar number of parameters. For instance, LDA-AQU surpasses DySample by 0.7 PQ (42.2 PQ v.s 41.5 PQ). Even with a strong backbone like ResNet-101, LDA-AQU still achieves a PQ gain of 1.5 (43.7 PQ v.s 42.2 PQ), surpassing DySample by 0.7 PQ (43.7 PQ v.s 43.0 PQ).

\begin{table}[!tbp]
  \caption{Performance comparison of Panoptic Segmentation on MS COCO based on Panoptic FPN.}
  \centering
  \begin{adjustbox}{width=0.45\textwidth}
  \begin{tabular}{l|c|ccccc|c}
  \toprule
  Panoptic FPN & Backbone & $PQ$ & $PQ^{th}$ & $PQ^{st}$ & $SQ$ & $RQ$ & Params \\
  \midrule 
  Nearest & ResNet-50 & 40.2 & 47.8 & 28.9 & 77.8 & 49.3 & 46.0M \\
  Deconv & ResNet-50 & 39.6 & 47.0 & 28.4 & 77.1 & 48.5 & +1.8M \\
  PS~\cite{shi2016real} & ResNet-50 & 40.0 & 47.4 & 28.8 & 77.1 & 49.1 & +7.1M \\
  CARAFE~\cite{wang2019carafe} & ResNet-50 & 40.8 & 47.7 & 30.4 & 78.2 & 50.0 & +0.2M \\
  IndexNet~\cite{lu2019indices} & ResNet-50 & 40.2 & 47.6 & 28.9 & 77.1 & 49.3 & +6.3M \\
  A2U~\cite{dai2021learning} & ResNet-50 & 40.1 & 47.6 & 28.7 & 77.3 & 48.0 & +29.2K \\
  FADE~\cite{lu2022fade} & ResNet-50 & 40.9 & 48.0 & 30.3 & 78.1 & 50.1 & +0.1M \\
  SAPA-B~\cite{lu2022sapa} & ResNet-50 & 40.6 & 47.7 & 29.8 & 78.0 & 49.6 & +0.1M \\
  DySample~\cite{liu2023learning} & ResNet-50 & \underline{41.5} & \underline{48.5} & \underline{30.8} & \underline{78.3} & \underline{50.7} & +49.2K\\
  \rowcolor{mygray} LDA-AQU & ResNet-50 & \textbf{42.2} & \textbf{48.7} & \textbf{32.4} & \textbf{78.6} & \textbf{51.5} & +0.1M \\
  \midrule
  Nearest & ResNet-101 & 42.2 & 50.1 & 30.3 & 78.3 & 51.4 & 65.0M \\
  CARAFE~\cite{wang2019carafe} & ResNet-101 & 42.8 & 49.7 & \underline{32.5} & \underline{79.1} & 52.1 & +0.2M \\
  DySample~\cite{liu2023learning} & ResNet-101 & \underline{43.0} & \underline{50.2} & 32.1 & 78.6 & \underline{52.4} & +49.2K \\
  \rowcolor{mygray} LDA-AQU & ResNet-101 & \textbf{43.7} & \textbf{50.3} & \textbf{33.5} & \textbf{79.6} & \textbf{53.0} & +0.1M \\
  \bottomrule
  \end{tabular}
  \end{adjustbox}
  \label{tab:panoptic_fpn}
\end{table}

\subsection{Semantic Segmentation}
As shown in Table~\ref{tab:upernet}, by replacing the upsamplers with LDA-AQU in the Feature Pyramid Network (FPN) and Multi-level Feature Fusion (FUSE) of UperNet, the mIoU of baseline model has been improved from $39.78$ to $42.31$, surpassing CARAFE by $1.31$ mIoU and DySample by $1.23$ mIoU. In addition, we also report the aAcc and mAcc metrics of the model. As depicted in Table \ref{tab:upernet}, the performance of LDA-AQU remains superior on these two metrics compared to other upsamplers.

\begin{table}[!t]
  \caption{Performance comparison of Semantic Segmentation on ADE20K based on UperNet. }
    \centering
    \begin{tabular}{l|c|ccc}
    \toprule
    UperNet & Backbone & aAcc & mIoU & mAcc \\
    \midrule
    Bilinear & ResNet-50 & 79.08 & 39.78 & 52.81 \\
    PS~\cite{shi2016real} & ResNet-50 & 79.34 & 39.10 & 50.36 \\
    CARAFE~\cite{wang2019carafe} & ResNet-50 & 79.45 & 41.0 & 52.59 \\
    FADE~\cite{lu2022fade} & ResNet-50 & \underline{79.97} & \underline{41.89} & \underline{54.64} \\
    SAPA-B~\cite{lu2022sapa} & ResNet-50 & 79.47 & 41.08 & 53.67 \\
    DySample~\cite{liu2023learning} & ResNet-50 & 79.82 & 41.08 & 53.00 \\
    \rowcolor{mygray} LDA-AQU & ResNet-50 & \textbf{80.11} & \textbf{42.31} & \textbf{55.37} \\
    \midrule
    Bilinear & ResNet-101 & 80.27 & \underline{42.52} & 54.91 \\
    DySample~\cite{liu2023learning} & ResNet-101 & \underline{80.30} & 42.39 & \underline{55.82} \\
    \rowcolor{mygray} LDA-AQU & ResNet-101 & \textbf{80.52} & \textbf{43.41} & \textbf{56.53} \\
    \bottomrule
    \end{tabular}
    \label{tab:upernet}
\end{table}

\subsection{Ablation Study} \label{sec:ablation}
We conduct ablation studies on MS COCO using Faster R-CNN and Mask R-CNN to verify the impact of hyperparameters in LDA-AQU.

\textbf{Local Deformation Ranges. }
Initially, we assess the impact of the hyperparameter $\theta$ on both object detection and instance segmentation tasks. As shown in Table~\ref{tab:abla_range_detection}, when the $\theta$ of each LDA-AQU in FPN is set to $11$, the model achieves optimal performance (\ie 39.2 AP). As $\theta$ decreases, the performance of the model gradually diminishes. We believe the reason is that a few neighboring points are enough to cover sufficient information to ensure accurate interpolation. Therefore, the model prefers to find broader contextual information as auxiliary items to optimize the upsampling results.

\begin{table}[!bpt]
	\caption{Performance comparison of various local deformation ranges in the FPN of Faster R-CNN.}
	\centering
  \begin{tabular}{c|ccc|ccc}
    \toprule
    $\theta$ & $AP$ & $AP_{50}$ & $AP_{75}$ & $AP_S$ & $AP_M$ & $AP_{L}$ \\
    \midrule
    5 & 38.7 & 60.3 & 42.0 & \textbf{23.0} & 42.3 & 49.8 \\
    7 & 39.0 & 60.5 & 42.5 & 22.9 & 42.6 & 50.1 \\
    9 & 39.1 & 60.6 & 42.3 & \textbf{23.0} & \textbf{43.0} & 49.9 \\
    11 & \textbf{39.2} & \textbf{60.7} & \textbf{42.7} & 22.9 & \textbf{43.0} & 50.1 \\
    13 & 39.0 & 60.4 & 42.4 & 22.8 & \textbf{43.0} & \textbf{50.3} \\
    \bottomrule
    \end{tabular}
		\label{tab:abla_range_detection}
\end{table}

\begin{figure*}[!thbp]
  \includegraphics[width=\linewidth]{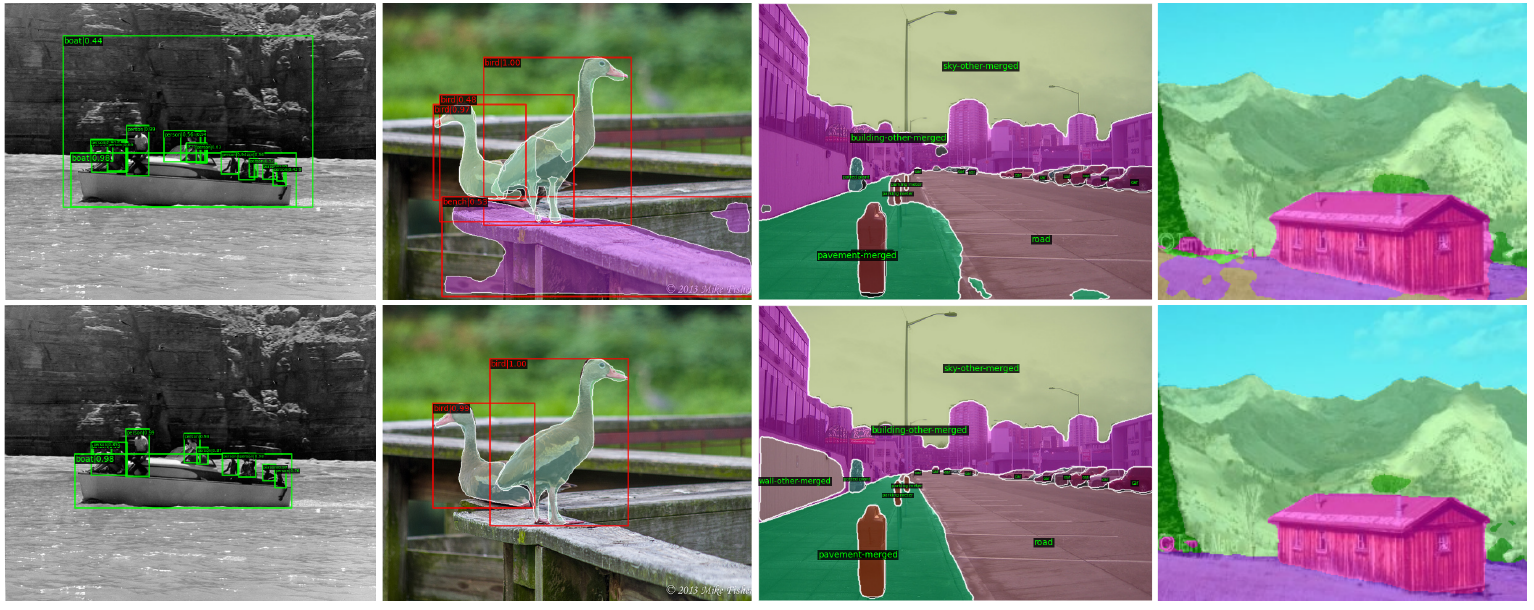}
  \caption{Qualitative comparison between baseline models (first row) and LDA-AQU (second row) across various tasks (\ie object detection, instance segmentation, panoptic segmentation, and semantic segmentation, from left to right).}
  \label{fig:visual_results}
  \vspace{-0.3cm}
\end{figure*}
Then, we verify the influence of $\theta$ in the mask head of Mask R-CNN. As shown in Table~\ref{tab:abla_range_mask}, when $\theta$ of mask head is set to $5$, the model achieves the best performance. The reason is that the size of the input feature map of the mask head is $14\times14$, so using a larger $\theta$ will make it difficult for the model to focus on local details.

\begin{table}[!ht]
	\caption{Performance comparison of various local deformation ranges in the mask head of Mask R-CNN. The best performance in detection is highlighted in bold, while the best in segmentation is underlined.}
	\centering
  \begin{tabular}{c|c|ccc|ccc}
    \toprule
    $\theta$ & Task & $AP$ & $AP_{50}$ & $AP_{75}$ & $AP_S$ & $AP_M$ & $AP_{L}$ \\
    \midrule
    \multirow{2}{*}{3} & bbox & 39.6 & 60.6 & 43.2 & 23.7 & 42.9 & 51.1 \\
    & segm & 36.0 & 57.6 & 38.4 & 17.3 & 38.4 & 52.7 \\
    \midrule
    \multirow{2}{*}{5} & bbox & \textbf{39.7} & \textbf{60.8} & \textbf{43.5} & \textbf{23.8} & \textbf{43.6} & \textbf{51.2} \\
    & segm & \underline{36.2} & \underline{57.9} & 38.5 & 17.3 & 39.1 & \underline{52.8} \\
    \midrule
    \multirow{2}{*}{7} & bbox & 39.4 & 60.5 & 43.0 & 23.1 & 43.2 & 50.8 \\
    & segm & 36.0 & 57.4 & 38.5 & 17.1 & 38.8 & 52.4 \\
    \bottomrule
    \end{tabular}
		\label{tab:abla_range_mask}
\end{table}

\textbf{Offset Groups. }
We evaluate the impact of the offset groups on model performance. As shown in Table~\ref{tab:abla_groups}, the model achieves the optimal performance when the number of groups is set to $2$. Excessive groupings will lead to a decreased size of features utilized for predicting offsets, thus impeding model learning.

\begin{table}[!ht]
	\caption{Performance comparison of different offset groups.}
	\centering
  \begin{tabular}{c|ccc|ccc}
    \toprule
    Groups & $AP$ & $AP_{50}$ & $AP_{75}$ & $AP_S$ & $AP_M$ & $AP_{L}$ \\
    \midrule
    1 & 38.9 & 60.3 & 42.1 & 22.8 & 42.5 & \textbf{50.3} \\
    2 & \textbf{39.2} & \textbf{60.7} & \textbf{42.7} & 22.9 & \textbf{43.0} & 50.1 \\
    4 & 39.0 & 60.3 & 42.3 & 23.0 & 42.7 & \textbf{50.3} \\
    8 & 38.8 & 60.2 & 42.1 & \textbf{23.1} & 42.6 & 50.1 \\
    \bottomrule
    \end{tabular}
		\label{tab:abla_groups}
\end{table}

\textbf{Channel Size Reduction Factors. }
Finally, we evaluate the impact of channel reduction factors on model performance. As illustrated in Table~\ref{tab:abla_channel}, when the reduction factor is set to $16$, LDA-AQU yields an AP gain of $1.4$ for Faster R-CNN (38.9 AP v.s 37.5 AP). By reducing the channel reduction factor, the performance of the model gradually improved, finally reaching $39.4$ AP. In order to balance the performance and computational complexity of the model, we set the channel reduction factor of LDA-AQU to $4$.

\begin{table}[!ht]
	\caption{Performance comparison of various channel size reduction factors.}
	\centering
  \begin{tabular}{c|ccc|ccc}
    \toprule
    Factor & $AP$ & $AP_{50}$ & $AP_{75}$ & $AP_S$ & $AP_M$ & $AP_{L}$\\
    \midrule
    16 & 38.9 & 60.4 & 42.4 & 23.3 & 42.8 & 49.7\\
    8 & 38.9 & 60.3 & 42.2 & 23.3 & 42.8 & 49.7\\
    4 & 39.2 & 60.7 & 42.7 & 22.9 & \textbf{43.0} & 50.1\\
    2 & \textbf{39.4} & \textbf{60.8} & \textbf{42.8} & \textbf{23.7} & \textbf{43.0} & \textbf{50.4} \\
    \bottomrule
    \end{tabular}
		\label{tab:abla_channel}
\end{table}
\subsection{Visual Inspection and Analysis}
In this section, we provide some visualizations and analysis of deformed neighbor points and results across different tasks. More visualizations can be found in the supplementary materials.

\textbf{Deformed Neighboring Points. }
We visualize the locations of some upsampled points and their corresponding deformed neighboring points. As shown in Figure~\ref{fig:visual_pts}, LDA-AQU can adaptively adjust the position of neighbor points according to the query features (\eg the fork in the fourth column). 
Even in occlusion scenes (\eg third column of Figure~\ref{fig:visual_pts}), LDA-AQU demonstrates a tendency to prioritize the featrues of object itself over the occluder (\ie branches). 

\textbf{Qualitative Experiments. }
We also conduct qualitative experiments to verify the effectiveness of LDA-AQU. 
Specifically, we visualize the the results of the baseline models and LDA-AQU across various visual tasks. As shown in Figure~\ref{fig:visual_results}, our model exhibits superior performance, thereby validating the effectiveness of LDA-AQU. 

\section{Conclusion}
In this paper, we introduce local self-attention into the upsampling task. Compared with previous methods, the upsampling based on local self-attention (LA-AQU) naturally incorporates the feature guidance mechanism without necessitating high-resolution input. Additionally, to enhance the adaptability of LA-AQU to complex upsampling scenarios, we further introduce the query-guided deformation mechanism and propose LDA-AQU. Finally, LDA-AQU can dynamically adjust the location and aggregation weight of neighboring points based on the features of the upsampled point. Through extensive experiments on four dense prediction tasks, we evaluate the effectiveness of LDA-AQU. Specifically, LDA-AQU has consistently demonstrated leading performance across above tasks, while maintaining a comparable FLOPs and parameters. For future work, we intend to explore dynamic local deformation ranges and investigate additional application scenarios, including image restoration, image inpainting, downsampling, \etc

\bibliographystyle{ACM-Reference-Format}
\bibliography{sample-base}

\appendix
\section{More Experimental Details}
We evaluate the effectiveness of LDA-AQU across four dense prediction tasks (\ie object detection, instance segmentaion, panoptic segmentaion, and semantic segmentaion) by substituting the base upsampler (\ie Nearest Neighbor Upsampler or Bilinear Upsampler) with LDA-AQU. Following the same process, we further compare the performance of LDA-AQU with other state-of-the-art upsamplers, including CARAFE~\cite{wang2019carafe}, IndexNet~\cite{lu2019indices}, FADE~\cite{lu2022fade}, SAPA~\cite{lu2022sapa}, DySample~\cite{liu2023learning}, \etc

\textbf{Object Detection. }
We utilize Faster R-CNN~\cite{ren2015faster} with ResNet~\cite{he2016deep} as the baseline model to conduct the performance comparison between various upsamplers and LDA-AQU on object detection task using the MS COCO dataset~\cite{lin2014microsoft}. Specifically, we compare the performance of various upsamplers by replacing the upsampler utilized in the Feature Pyramid Network (FPN)~\cite{lin2017feature}. During both the training and testing stages, the short side of input images is resized to $800$ pixels. We perform performance comparisons and ablation studies on 4 GPUs with 2 images per GPU. The SGD optimizer is utilized to train the network with a momentum of 0.9 and a weight decay of 0.0001. Following previous work~\cite{goyal2017accurate}, we set the initial learning rate to $0.01$. We employ the 1$\times$ (12 epochs) training schedule to train networks, with the learning rate decreasing by a factor of 0.1 at the 9-$th$ and 11-$th$ epochs.

\textbf{Instance Segmentation. }
We utilize Mask R-CNN~\cite{he2017mask} with ResNet~\cite{he2016deep} as the baseline model to conduct the performance comparison on instance segmentaion task using the MS COCO dataset~\cite{lin2014microsoft}. We perform performance comparison by replacing the upsamplers of FPN and mask head in Mask R-CNN. All other settings remain the same as for object detection.

\textbf{Panoptic Segmentation. }
For panoptic segmentation task, we use Panoptic FPN~\cite{kirillov2019panopticFPN} with ResNet~\cite{he2016deep} as the baseline model to conduct the performance comparison using the MS COCO dataset~\cite{lin2014microsoft}. We perform performance comparison by replacing the upsamplers of FPN in Panoptic FPN. All other settings remain the same as for object detection.

\textbf{Semantic Segmentation. }
For semantic segmentation task, we utilize UperNet~\cite{xiao2018unified} with ResNet~\cite{he2016deep} as the baseline model to conduct the performance comparison using the ADE20K dataset~\cite{zhou2017scene}. We train the model on 4 GPUs with 2 images per GPU. We utilize the SGD optimizer with an initial learning rate of $0.005$, a momentum of 0.9, and a weight decay of 0.0005. The models are trained for 160K iterations, utilizing the \textit{Poly} learning rate policy with a \textit{power} of 0.9 and a \textit{min}\_\textit{lr} of 0.0001.

\section{More Ablation Studies}
As in the main text, all ablation studies are conducted on models based on Faster R-CNN~\cite{ren2015faster} and ResNet-50~\cite{he2016deep} using the MS COCO dataset~\cite{lin2014microsoft}.

\textbf{Kernel Sizes. }
We verify the impact of different kernel sizes of information encoding $k_{e}$ and neighboring points $k_{u}$ in LDA-AQU on model performance. As illustrated in Table~\ref{tab:abla_ks_detection}, the impact of $k_{e}$ and $k_{u}$ on model performance is not as significant as that of the local deformation ranges. Larger values of $k_{e}$ and $k_{u}$ typically result in greater performance improvements, but they also lead to a notable increase in computational complexity and parameters. Therefore, we prefer to set $k_{e}$ and $k_{u}$ to a kernel size of $3\times 3$ to achieve a better balance between computational complexity and performance.

\begin{table}[!h]
	\caption{Performance comparison of various kernel sizes in the FPN of Faster R-CNN.}
	\centering
  \begin{tabular}{cc|ccc|ccc}
    \toprule
    $k_{e}$ & $k_{u}$ & $AP$ & $AP_{50}$ & $AP_{75}$ & $AP_S$ & $AP_M$ & $AP_{L}$ \\
    \midrule
    1 & 3 & 38.9 & 60.5 & 42.3 & 22.8 & 42.4 & 50.1 \\
    1 & 5 & 39.1 & 60.4 & 42.8 & 23.3 & 43.3 & 49.9 \\
    3 & 3 & 39.2 & 60.7 & 42.7 & 22.9 & 43.0 & 50.1 \\
    3 & 5 & 39.2 & 60.5 & 42.6 & 23.2 & 42.9 & 50.2 \\
    5 & 3 & 39.3 & 60.8 & 42.7 & 23.3 & 43.3 & 50.3 \\
    \bottomrule
    \end{tabular}
		\label{tab:abla_ks_detection}
\end{table}

\textbf{Upsampling scheme. }
We also evaluate the impact of different upsampling methods for the queries in LDA-AQU on model performance.
As shown in Table~\ref{tab:abla_up_detection}, the nearest neighbor interpolation upsampler performs the same as the bilinear upsampler, with an AP of 39.2. However, their detection performance on objects of different scales differs significantly. 
We posit that, due to the higher resolution of feature maps utilized for detecting small objects, employing nearest neighbor points as precise object features to guide the upsampling task is more appropriate. In contrast, for low-resolution feature maps with a notable semantic gap between points, bilinear interpolated features are preferred for guiding the upsampling task due to their greater resemblance to the features of the objects.

\begin{table}[!h]
	\caption{Performance comparison of different upsampling methods for query upsampling in LDA-AQU.}
	\centering
  \begin{tabular}{c|ccc|ccc}
    \toprule
    Method & $AP$ & $AP_{50}$ & $AP_{75}$ & $AP_S$ & $AP_M$ & $AP_{L}$ \\
    \midrule
    Nearest & 39.2 & 60.6 & 42.3 & 23.8 & 42.7 & 50.3 \\
    Bilinear & 39.2 & 60.7 & 42.7 & 22.9 & 43.0 & 50.1 \\
    \bottomrule
    \end{tabular}
		\label{tab:abla_up_detection}
\end{table}

\begin{table*}[!ht] 
  \caption{Performance comparison of Object Detection on Pascal VOC based on Faster R-CNN (F-RCNN). The best in each column is highlighted in bold, and the second best is underlined.}
      \resizebox{\textwidth}{!}{
      \setlength
      \tabcolsep{2pt}
      \begin{tabular}{lc|@{}c|@{}c|@{}c|@{}c|@{}c|@{}c|@{}c|@{}c|@{}c|@{}c|@{}c|@{}c|@{}c|@{}c|@{}c|@{}c|@{}c|@{}c|@{}c|@{}c}
          \hline
          F-RCNN~\cite{ren2015faster} & \rotatebox{0}{\textbf{mAP}} & \rotatebox{60}{plane} & \rotatebox{60}{bicycle}	& \rotatebox{60}{bird} & \rotatebox{60}{boat} & \rotatebox{60}{bottle} & \rotatebox{60}{bus} & \rotatebox{60}{car} & \rotatebox{60}{cat} & \rotatebox{60}{chair} & \rotatebox{60}{cow}	& \rotatebox{60}{table} & \rotatebox{60}{dog}	& \rotatebox{60}{horse}	& \rotatebox{60}{bike} & \rotatebox{60}{person} & \rotatebox{60}{plant} & \rotatebox{60}{sheep} & \rotatebox{60}{sofa} & \rotatebox{60}{train} & \rotatebox{60}{tv} \\
          \hline
          Nearest & 78.7 & 80.5 & \textbf{87.5} & 78.1 & 67.2 & 66.3 & 86.1 & 87.5 & \underline{89.2} & 63.5 & 84.6 & 74.1 & 87.8 & 87.0 & 80.8 & 80.1 & 52.3 & 78.5 & 80.2 & 85.3 & 76.5 \\
          Deconv & 77.7 & 80.7 & 80.7 & 79.3 & 66.8 & 64.9 & \underline{86.3} & 87.1 & 88.5 & 61.8 & 84.4 & 74.6 & 87.0 & 85.8 & 79.7 & 79.8 & 51.8 & 77.1 & 79.7 & \textbf{86.4} & 71.1 \\
          PS~\cite{shi2016real} & 78.5 & 81.0 & 81.0 & 78.4 & 66.4 & 65.9 & 85.4 & 87.4 & 89.1 & 62.2 & 86.3 & 74.5 & 87.5 & 87.8 & 79.9 & 80.0 & 51.6 & \underline{84.0} & 79.5 & 85.2 & 77.5 \\
          CARAFE~\cite{wang2019carafe} & 79.2 & 81.1 & 81.2 & 79.2 & \textbf{73.3} & 65.8 & \underline{86.3} & \underline{88.1} & 89.0 & 63.0 & 84.8 & 73.4 & 88.2 & 88.0 & 80.2 & \underline{80.2} & \underline{53.2} & \textbf{84.8} & 81.3 & \underline{86.1} & 76.3 \\
          FADE~\cite{lu2022fade} & 79.3 & 81.5 & 81.1 & 79.2 & 68.1 & \underline{66.4} & 85.8 & \underline{88.1} & 89.1 & \underline{63.9} & \underline{86.4} & 75.4 & 87.5 & 88.4 & \textbf{87.0} & \underline{80.2} & \textbf{53.9} & 79.7 & \underline{82.4} & \underline{86.1} & 76.4 \\
          SAPA-B~\cite{lu2022sapa} & \underline{79.9} & \underline{87.0} & \underline{86.8} & \underline{79.7} & \underline{72.9} & 65.9 & 85.9 & 87.8 & 88.7 & 63.4 & \textbf{86.6} & 75.1 & \textbf{88.9} & \textbf{89.0} & 86.2 & 80.0 & 51.9 & 78.9 & 80.0 & 85.5 & \underline{77.9} \\
          DySample~\cite{liu2023learning} & 78.9 & 80.9 & 81.0 & \textbf{80.1} & 66.3 & 66.3 & \textbf{86.8} & 87.9 & \textbf{89.3} & \textbf{64.4} & 86.3 & \underline{75.9} & 87.9 & 88.4 & 86.5 & \underline{80.2} & 51.4 & 79.3 & \underline{82.0} & 79.8 & 77.4 \\
          LDA-AQU (ours) & \textbf{80.3} & \textbf{88.8} & 81.0 & 79.1 & 72.7 & \textbf{67.3} & \textbf{86.8} & \textbf{88.3} & 88.7 & 63.5 & \underline{86.4} & \textbf{79.2} & \underline{88.7} & \underline{88.6} & \underline{86.8} & \textbf{80.4} & \underline{53.2} & 79.8 & \textbf{82.6} & 85.9 & \textbf{78.4}\\
          \hline
      \end{tabular}
      }
      \label{tab:pascal_results}
  \end{table*}
\section{Object Detection on Pascal VOC}
We further evaluate the effectiveness of LDA-AQU on Pascal VOC dataset~\cite{everingham2010pascal}. Specifically, we utilize the VOC 2012 and VOC 2007 trainval splits for model training, and evaluate their performance across different models on the VOC 2007 test split. We resize both the training and testing images to $640\times 640$, ensuring consistency with the training strategy and hyperparameters employed in the experiments conducted on MS COCO. Similarly, we employ Faster R-CNN~\cite{ren2015faster} with nearest neighbor interpolation as the baseline model and compare the effects of different upsampling methods by modifying the upsamplers implemented in FPN. 

\textbf{Comparision with Other State-of-the-Art Upsamplers. }
As illustrated in Table~\ref{tab:pascal_results}, our LDA-AQU achieves the best performance with an AP of 80.3, surpassing the baseline model by 1.6 AP (80.3 AP vs. 78.7 AP).

\begin{figure}[!t]
  \includegraphics[width=1\linewidth]{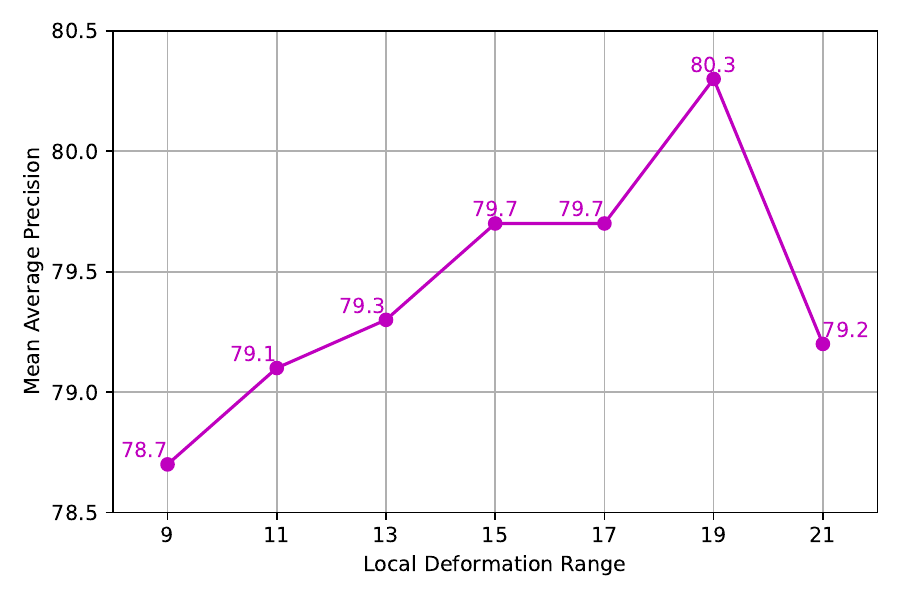}
  \caption{Performance comparison of different local deformation ranges in LDA-AQU on the Pascal VOC dataset.}
  \label{fig:sup_range}
  \vspace{-0.3cm}
\end{figure}

\textbf{Ablation Study of Local Deformation Ranges. }
Then, we evaluate the impact of varying the local deformation ranges in LDA-AQU on the Pascal VOC dataset. As depicted in Figure~\ref{fig:sup_range}, the model achieves optimal performance when $\theta$ is set to 19. This is attributed to the larger object scale typically in the Pascal VOC dataset compared to the MS COCO dataset, necessitating a higher value of $\theta$ to achieve better shape matching with the objects.

\section{More Visual Inspection and Analysis}
We augment our study with additional visualizations, including illustrations of deformed neighboring points, alongside qualitative experiments conducted across the aforementioned four tasks. All visualizations are based on the backbone network of ResNet-50~\cite{he2016deep}.

\begin{figure*}[!t]
  \includegraphics[width=0.85\linewidth]{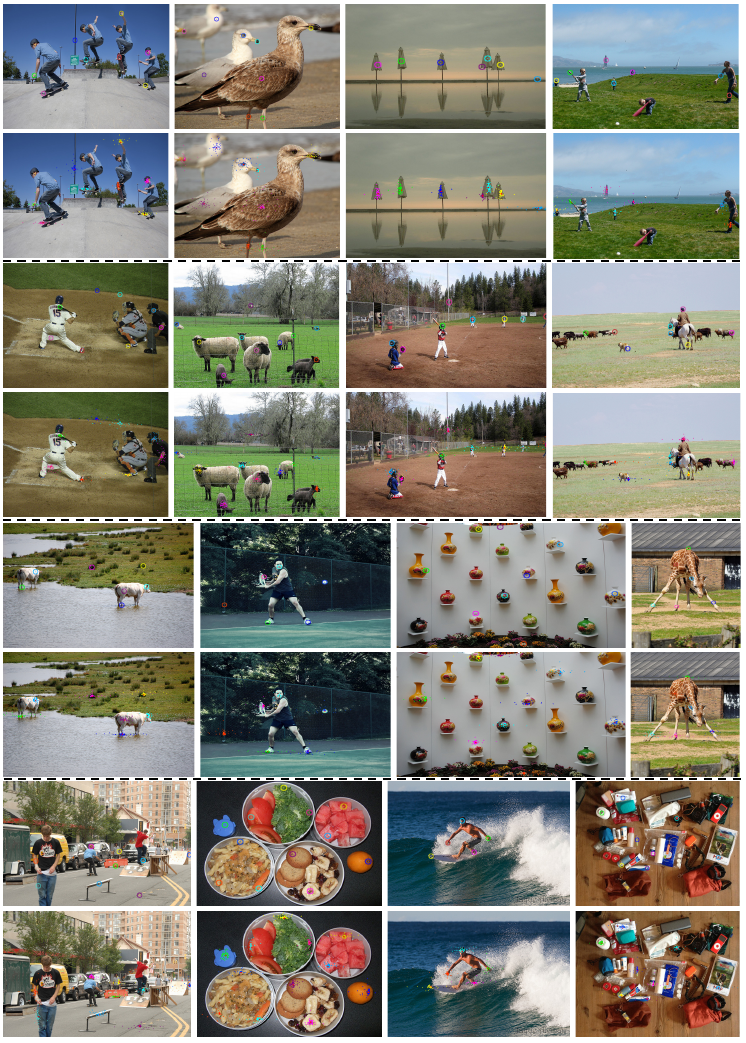}
  \caption{
    Visualization of upsampled points (colored rings) and their corresponding deformed neighboring points (scatters with the same color as the rings). Different groups of images are separated by black dotted lines, with the first row of each group representing the upsampled points and the second row representing the corresponding deformed neighboring points.
  }
  \label{fig:sup_pts}
  \vspace{-0.3cm}
\end{figure*}

\textbf{Deformed Neighboring Points. }
As shown in Figure~\ref{fig:sup_pts}, we visualize more upsampled points (\ie queries) and their corresponding deformed neighboring points. Many examples in Figure~\ref{fig:sup_pts} demonstrate that our LDA-AQU can adaptively adjust the position of neighboring points according to the shape (\eg umbrella, pole, leg, \etc) and context (\eg boundary of two different objects) of the objects, thereby focusing more on features related to the queries.

\begin{figure*}[!t]
  \includegraphics[width=0.85\linewidth]{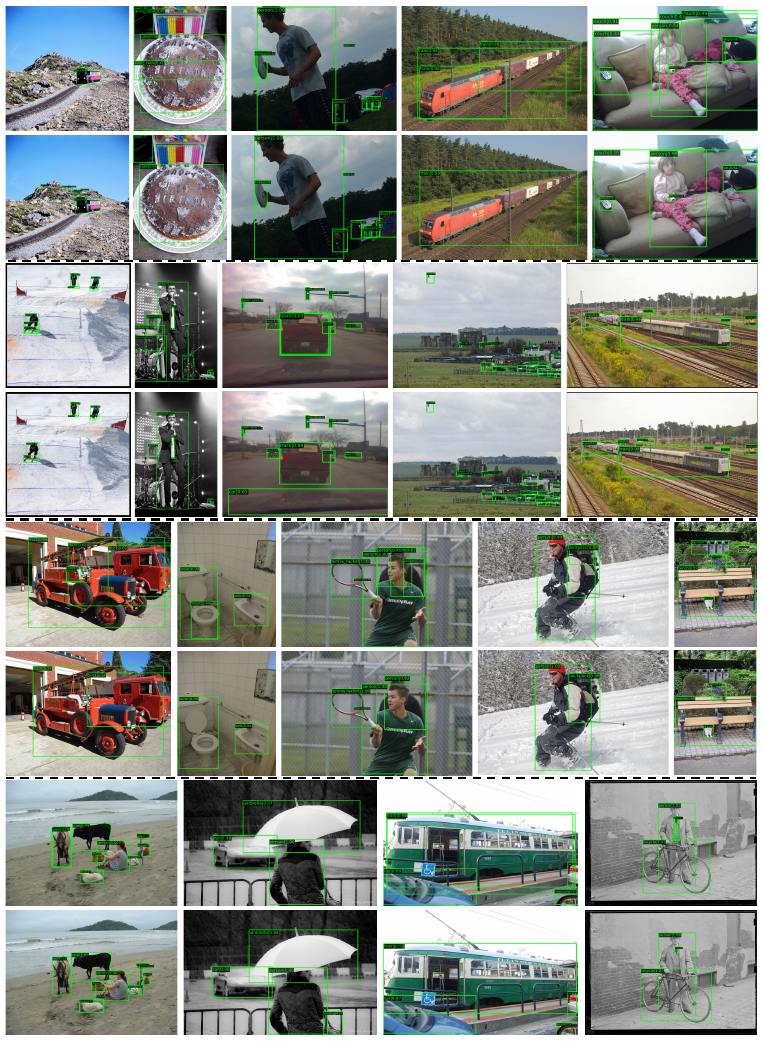}
  \caption{Visualization of prediction results based on Faster R-CNN on MS COCO. Different groups of images are separated by black dotted lines, with the first row of each group representing the results of Faster R-CNN w/ BI and the second row representing the results of Faster R-CNN w/ LDA-AQU.}
  \label{fig:sup_obj}
  \vspace{-0.3cm}
\end{figure*}

\textbf{Qualitative Experiments on Object Detection. }
As depicted in Figure~\ref{fig:sup_obj}, we visualize more object detection results to compare the effects of Faster R-CNN with Bilinear Interpolation (BI) and Faster R-CNN with LDA-AQU.

\begin{figure*}[!t]
  \includegraphics[width=0.85\linewidth]{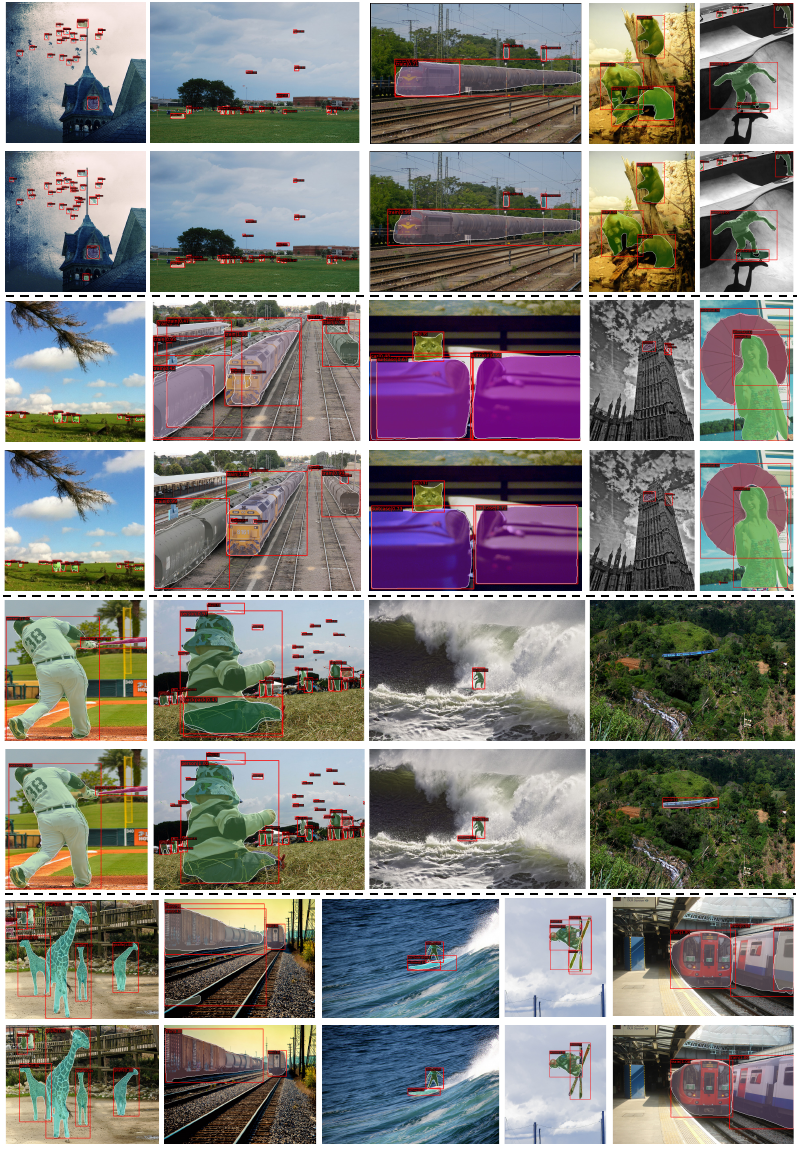}
  \caption{Visualization of prediction results based on Mask R-CNN on MS COCO. Different groups of images are separated by black dotted lines, with the first row of each group representing the results of Mask R-CNN w/ BI and the second row representing the results of Mask R-CNN w/ LDA-AQU.}
  \label{fig:sup_ins}
  \vspace{-0.3cm}
\end{figure*}

\textbf{Qualitative Experiments on Instance Segmentation. }
As illustrated in Figure~\ref{fig:sup_ins}, we present additional instance segmentation results to compare the performance of Mask R-CNN with BI and Mask R-CNN with LDA-AQU.

\begin{figure*}[!t]
  \includegraphics[width=0.85\linewidth]{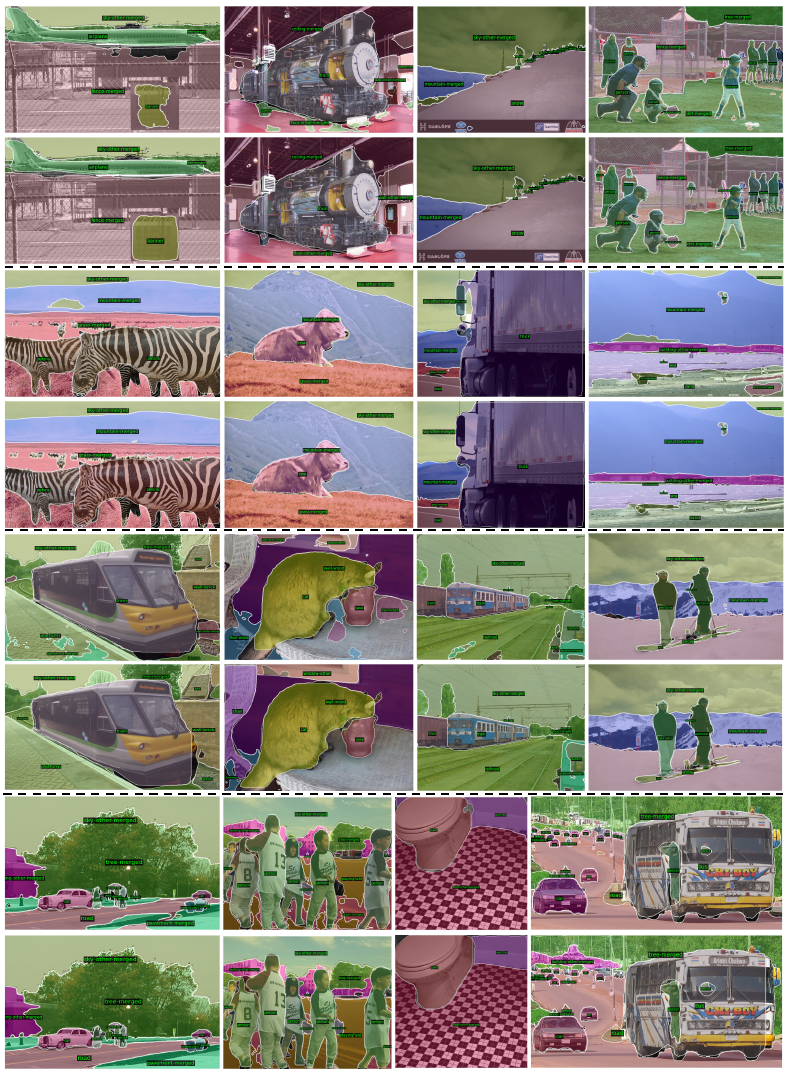}
  \caption{Visualization of prediction results based on Panoptic FPN on MS COCO. Different groups of images are separated by black dotted lines, with the first row of each group representing the results of Panoptic FPN w/ BI and the second row representing the results of Panoptic FPN w/ LDA-AQU.}
  \label{fig:sup_pan}
  \vspace{-0.3cm}
\end{figure*}

\textbf{Qualitative Experiments on Panoptic Segmentation. }
In Figure~\ref{fig:sup_pan}, we present additional panoptic segmentation results, comparing the effectiveness of Panoptic FPN with BI and Panoptic FPN with LDA-AQU.

\begin{figure*}[!t]
  \includegraphics[width=0.85\linewidth]{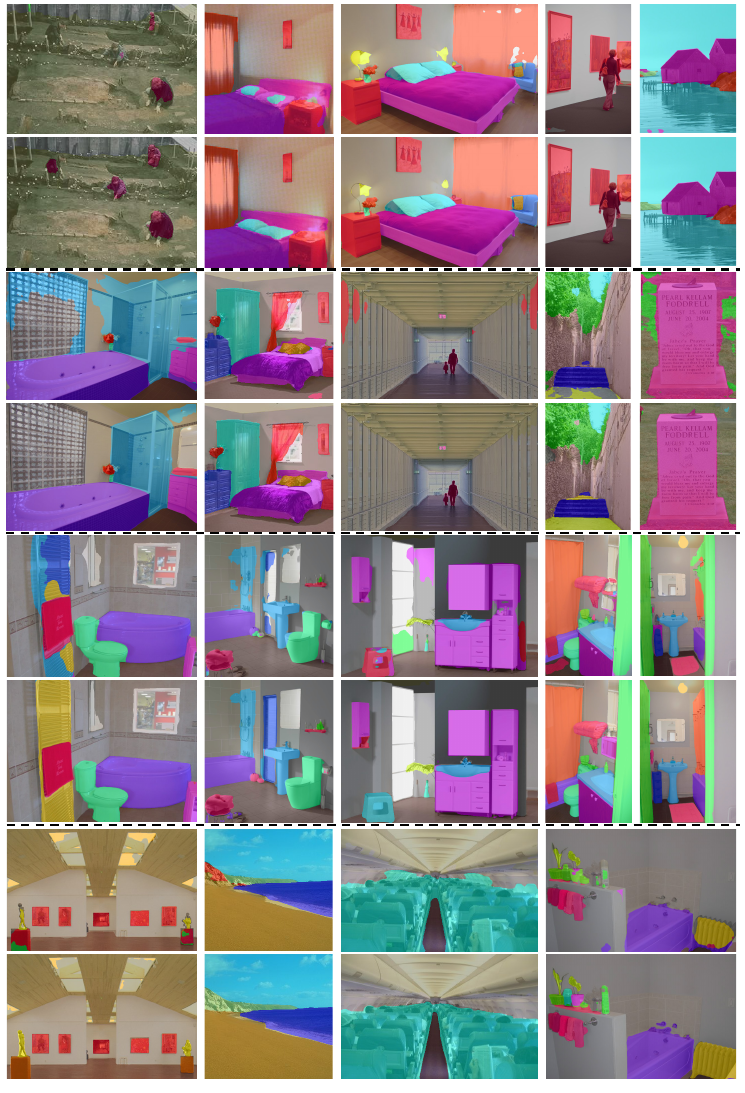}
  \caption{Visualization of prediction results based on UperNet on ADE20K. Different groups of images are separated by black dotted lines, with the first row of each group representing the results of UperNet w/ BI and the second row representing the results of UperNet w/ LDA-AQU.}
  \label{fig:sup_sem}
  \vspace{-0.3cm}
\end{figure*}

\textbf{Qualitative Experiments on Semantic Segmentation. }
In Figure~\ref{fig:sup_sem}, we present more semantic segmentation results, comparing the effectiveness of UperNet with BI and UperNet with LDA-AQU.

\end{document}